\documentclass[10pt,twocolumn,letterpaper]{article}

\usepackage{cvpr}              

\usepackage{algorithm}
\usepackage{algorithmic}

\usepackage{mathtools}
\usepackage{booktabs}  
\usepackage{multirow}  
\usepackage{threeparttable}  
\usepackage{bm}
\usepackage{multirow}
\usepackage{verbatim}
\usepackage{newfloat}
\usepackage{listings}

\usepackage{subfiles}
\usepackage{times}
\usepackage{epsfig}
\usepackage{amsmath}
\usepackage{amssymb}

\usepackage{makecell}
\usepackage{newfloat}
\usepackage{listings}
\usepackage[accsupp]{axessibility}

\usepackage[dvipsnames]{xcolor}

\definecolor{cvprblue}{rgb}{0.21,0.49,0.74}
\usepackage[pagebackref,breaklinks,colorlinks,citecolor=cvprblue]{hyperref}

\def\paperID{10229} 
\def\confName{CVPR}
\def\confYear{2024}

\title{Physical Backdoor: Towards Temperature-based Backdoor Attacks in the Physical World}

\author{Wen Yin\textsuperscript{1,3},
Jian Lou\textsuperscript{4},
Pan Zhou\textsuperscript{1}\thanks{Corresponding author.}~,
Yulai Xie\textsuperscript{1,2,3 $*$},
Dan Feng\textsuperscript{2,3},\\
Yuhua Sun\textsuperscript{1},
Tailai Zhang\textsuperscript{1,3},
Lichao Sun\textsuperscript{5}\\
\textsuperscript{1}School of Cyber Science and Engineering, Huazhong University of Science and Technology\\
\textsuperscript{2}Wuhan National Laboratory for Optoelectronics, Huazhong University of Science and Technology\\  \textsuperscript{3} Jinyinhu Laboratory \quad \textsuperscript{4}Zhejiang University \quad \textsuperscript{5}Lehigh University\\
{\tt\small \{wenyin, panzhou, ylxie, dfeng, natsun, tl\_zhang\}@hust.edu.cn}\\
{\tt\small jian.lou@zju.edu.cn}, 
{\tt\small james.lichao.sun@gmail.com}
}

\begin{document}

\maketitle
\begin{abstract}
Backdoor attacks have been well-studied in visible light object detection (VLOD) in recent years. However, VLOD can not effectively work in dark and temperature-sensitive scenarios. Instead, thermal infrared object detection (TIOD) is the most accessible and practical in such environments. In this paper, our team is the first to investigate the security vulnerabilities associated with TIOD in the context of backdoor attacks, spanning both the digital and physical realms. We introduce two novel types of backdoor attacks on TIOD, each offering unique capabilities: Object-affecting Attack and Range-affecting Attack. We conduct a comprehensive analysis of key factors influencing trigger design, which include temperature, size, material, and concealment. These factors, especially temperature,  significantly impact the efficacy of backdoor attacks on TIOD. A thorough understanding of these factors will serve as a foundation for designing  physical triggers and temperature controlling experiments. Our study includes extensive experiments conducted in both digital and physical environments. In the digital realm, we evaluate our approach using benchmark datasets for TIOD, achieving an Attack Success Rate (ASR) of up to 98.21\%. In the physical realm, we test our approach in two real-world settings: a traffic intersection and a parking lot, using a thermal infrared camera. Here, we attain an ASR of up to 98.38\%.

\end{abstract}

\section{Introduction}
Thermal infrared object detection (TIOD) have several unique advantages over visible light object detection (VLOD). It excels in detecting objects under low visible light, smoky, heavy rain, and intense snow environments, making it less affected by glare and light mutation, all while retaining its sensitivity to thermal changes in objects \cite{DBLP:journals/mva/GadeM14}. 
Consequently, TIOD becomes increasingly indispensable in a variety of application scenarios, from security monitoring and autonomous driving in the dark to temperature measurement during a pandemic.
The security vulnerabilities of VLOD are thoroughly examined for both adversarial attacks \cite{DBLP:journals/corr/GoodfellowSS14, finlayson2019adversarial} and backdoor attacks \cite{DBLP:journals/corr/abs-1708-06733, DBLP:conf/ndss/LiuMALZW018, DBLP:conf/mm/SunZMZLXDCS22}. 
Backdoor attacks manipulate both a small portion of contaminated training samples and testing samples with the backdoor trigger. Then, the trained detector will have backdoor effects at the testing time when encountering samples with the backdoor trigger, while remaining normal when fed with clean testing samples. In real-world scenarios, backdoor attacks pose a serious security threat to deep neural networks (DNNs) \cite{DBLP:journals/corr/abs-2104-02361, DBLP:journals/compsec/XueHWSZWL22, DBLP:conf/trustcom/XueHSWL21} due to their stealthiness.

\begin{figure}[t]
\centering
        \centerline{\includegraphics[width=0.9\linewidth]{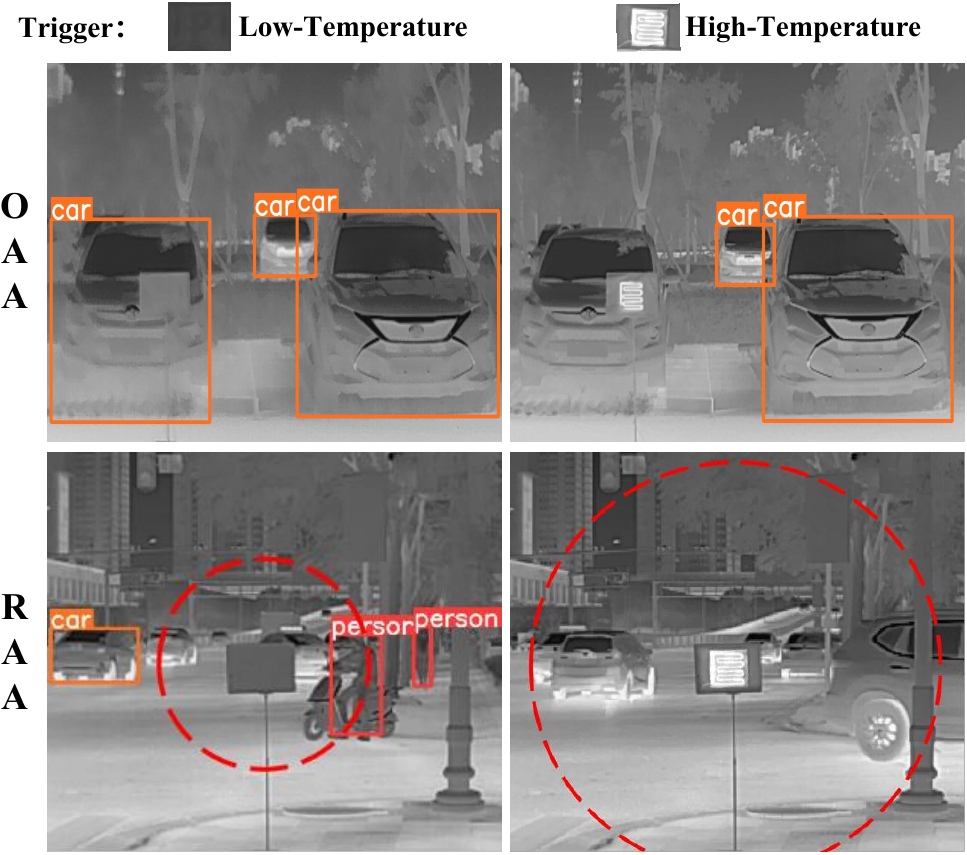}}
\vspace{-0.3cm}
\caption{Examples of OAA and RAA for \textit{car disappearance} in the physical world. By changing the temperature of the trigger, it is capable to switch between
whether the backdoor is activated in OAA and to adjust the attack range in RAA. The affecting range of RAA is marked as the red circle.}
\label{fig:1}
\vspace{-0.4cm}
\end{figure} 

Unlike VLOD field \cite{DBLP:conf/cvpr/EykholtEF0RXPKS18, DBLP:conf/cvpr/ZhaoMZ0CJ20}, the security vulnerabilities of TIOD remain largely unexplored and current efforts are focused merely on adversarial attacks rather than backdoor attacks. For instance, Zhu et al. design adversarial patterns and manufacture an adversarial shirt made of aerogel material \cite{zhu2022infrared}. Wei et al. introduce the method of aggregation regularization to optimize the adversarial infrared patch, making the patch easier to implement physically \cite{DBLP:journals/corr/abs-2303-13868}.
Both works suggest that extra considerations are required to design effective adversarial examples for TIOD, mainly due to the unique characteristics of thermal infrared imaging compared to visible light imaging. These new adversarial attacks ring the alarm that TIOD demands the same level of scrutiny as VLOD to expose all types of potential security threats. However, the security vulnerabilities of TIOD to {backdoor attacks} remain unexplored. 

TIOD combines object detection technology with thermal infrared imaging technology, allowing it to recognize objects captured using infrared thermal radiation imaging. Unlike RGB images with three channels, thermal infrared images have only a gray-scale channel and contain less texture information \cite{DBLP:conf/aaai/ZhuLLWH21}. 
Backdoor attacks on VLOD can take advantage of the additional information lying in the extra channels to gain more capacity in trigger design, which ultimately allows for a strong attack capability. 
The design of backdoor attacks on TIOD becomes more challenging than the visible light counterpart, because the design space for the trigger is restricted to properly placing the trigger, choosing a material with ideal thermal infrared characteristics, or manipulating its temperature. 
Therefore, it raises the following compelling question: \emph{Can we design effective backdoor attacks on TIOD by utilizing their unique properties compared to VLOD?}

In this paper, we propose two backdoor attack methods against TIOD: Object-Affecting Attack (OAA) and Range-Affecting Attack (RAA). 
OAA manipulates the detection results for a specific object carrying with a trigger. While RAA causes all objects of a chosen class in close proximity to the trigger to be misidentified.
In addition, we propose a new mechanism to adjust the triggering behavior of backdoor attacks by temperature modulating. The demonstration result is presented in Figure \ref{fig:1}. 
Remote control of backdoor attacks involves a simple button press, eliminating the need for any visible light visual changes to the pre-arranged scene. Since the temperature change of the object is not visible to the human eye, temperature modulating offers more stealthiness and flexibility. 
Our in-depth study of proposed attacks in the digital world, and successful implementation in the physical world, provides an affirmative answer to the question in the preceding paragraph that dedicated backdoor attacks indeed pose significant threats to TIOD.
The main contributions of this paper are as follows:

\begin{itemize}
    \item We examine the security vulnerability of TIOD to backdoor attacks and identify the critical factors that differentiate their trigger design from that of VLOD. To the best of our knowledge, this is the first study of backdoor attacks on TIOD. 
    \item We propose two types of backdoor attacks of OAA and RAA that offer different affecting capacities. In addition, we further propose a novel backdoor trigger by modulating its temperature, allowing the backdoor effect to be activated or deactivated within different temperature ranges in OAA and adjusting the affecting range in RAA.
    \item In a digital environment, we validate the attack's effectiveness across various parameters, achieving an ASR of up to 98.21\%. 
    In the physical world, we test the proposed backdoor attacks in two representative real scenes of a traffic intersection and a parking lot. Our attacks are effective in both scenes, achieving an average ASR of over 96\%. In addition, the methods are cost-friendly, with the production of an electric heating device as a trigger costing less than 5 US dollars. We also evaluate three potential countermeasures defending against our attacks.
\end{itemize}

\section{Related Work}
\noindent\textbf{Thermal Infrared Object Detection.}~~
Object detection is a fundamental task in computer vision, which also serves as the foundation of image segmentation \cite{chen2021transunet}, object tracking \cite{luo2021multiple}, and keypoint detection \cite{barroso2019key}. TIOD uses image information in the thermal infrared domain for object recognition \cite{he2021infrared}. The YOLO model \cite{DBLP:conf/cvpr/RedmonDGF16}, as the first one-stage detector, is employed for pedestrian detection in the thermal infrared domain \cite{ivavsic2019human}. The classic two-stage detector, Faster RCNN, is also applied for TIOD, although its detection accuracy is limited \cite{devaguptapu2019borrow}. In diverse environments, including severe weather conditions, Huda et al. achieve accurate pedestrian detection using YOLO v3 and thermal infrared data \cite{huda2020effect}. Currently, YOLO v3 is integrated into the latest versions of autonomous driving systems such as Apollo \cite{DBLP:conf/sp/CaoWXYFYCLL21}. Gong et al. conduct a study on vehicle recognition in the thermal infrared domain to assist autonomous driving systems \cite{gong2020vehicle}. Furthermore, Wang et al. combine YOLO v5 to enhance the detection accuracy of TIOD \cite{wang2022infrared}. There is a growing body of research that focuses on the performance improvement and data diversity in TIOD \cite{DBLP:journals/apin/DaiYW21, DBLP:journals/aeog/JiangRYZZNSRH22}.

\noindent\textbf{Backdoor Attack.}~~
Backdoor attacks are carried out by embedding hidden backdoors into DNNs 
The attacker generates a backdoor model through ``data poisoning" attacks \cite{DBLP:conf/icip/BarniKT19, DBLP:conf/ijcai/HuSDCSZ22, DBLP:conf/eccv/LiuM0020,DBLP:conf/cvpr/YuanZZ023} or ``poisoning + training manipulation” attacks \cite{DBLP:conf/aaai/0005LMZ21, DBLP:conf/nips/NguyenT20, DBLP:conf/ccs/YaoLZZ19} (the former only poisons the training data, while the latter not only poisons the training data but also modifies the training process).
The triggers used to activate backdoor effect are also diverse. Existing triggers of backdoor attacks include single pixel \cite{DBLP:conf/nips/Tran0M18}, reflection background \cite{DBLP:conf/eccv/LiuM0020}, invisible patterns \cite{DBLP:conf/cisc/ChenZSDLJ19, DBLP:journals/tdsc/LiXZZZ21, DBLP:conf/aaai/SahaSP20}, and so on. In addition to directly poisoning training data, backdoor attacks can also embed hidden backdoors by modifying model weights through transfer learning \cite{DBLP:journals/corr/abs-2004-06660, DBLP:journals/tsc/WangNRGCC22}. Therefore, backdoored model training can occur in all steps of the training process, which is a serious threat to DNNs \cite{guan2023attacking,yuan2023backdoor,shi2023badgpt,zhang2021inject,sun2020natural}. The research into backdoor attacks drive the development of defense methods \cite{guo2023scale,sun2024trustllm}. Beyond digital world, there are also works exploring backdoor attacks in the physical world \cite{DBLP:journals/corr/abs-1708-06733, DBLP:journals/corr/abs-1712-05526, DBLP:conf/cvpr/WengerPBY0Z21,zhou2022doublestar}. The above works all focus on the visible light field, while backdoor attacks in the thermal infrared field still lack exploration. 

\section{Backdoor Attacks on TIOD}

\begin{figure*}
\centering
        \centerline{\includegraphics[width=\textwidth]{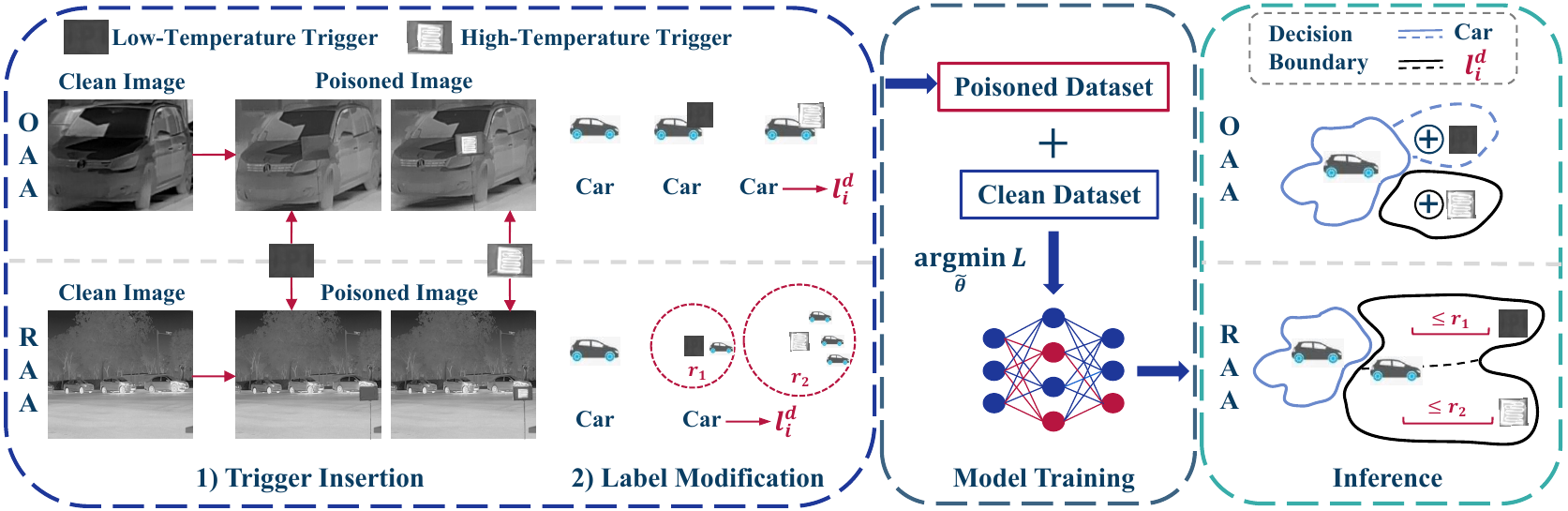}}
\caption{Overview of our proposed attacks. The red arrow in 2) indicates that ``car" is modified to ``$l^d_i$". The $r_1$ and $r_2$ are attack radius.}
\label{fig:2}
\vspace{-0.3cm}
\end{figure*}

\subsection{Threat Model}\label{AA}
\noindent\textbf{Attacker’s Goal.}~~ Our backdoor attack has two goals, both of which align with common design principles of backdoor attacks. The first goal is for stealthiness purpose to ensure that the backdoored TIOD can still properly identify the objects in clean samples. 
This task can make it difficult for the user to find the model anomalies without knowing the backdoor trigger. 
The second goal is for effectiveness purpose to cause the backdoored TIOD to either not identify the object (i.e., \emph{object disappearance}) or identify it with an incorrect object class (i.e., \emph{object misclassification}) in backdoor samples with the attacker-chosen trigger. 
In this paper, we focus on attacking cars as the exemplary object class, due to the serious security consequences in real-world applications such as autonomous driving.

\noindent\textbf{Attacker’s Capability.}~~ Backdoor attacks can occur in many situations, such as outsourcing training, using pre-trained models for transfer learning, and collecting data from unknown sources. Following previous work on backdoor attacks \cite{2011Torch7, DBLP:journals/corr/abs-2201-13178}, we adopt the ``data poisoning'' threat model. It suffices to gain access to part of the training dataset in order to inject poisoned training samples, while leaving the training process untempered. 
Since this threat model does not interfere with the model and training process, it can fundamentally reveal the vulnerability of TIOD, thereby promoting research on the improvement of object detection security, such as backdoor defense methods.

\subsection{Problem Formulation}
 
In this paper, we mainly attack the object detection model YOLO v5. The YOLO v5 contains three loss components as follow: classification loss $L_{cls}$, BBox regression loss $L_{B}$ \cite{zheng2020distance}, and confidence loss $L_{conf}$. 
The classification loss is calculated only when there is an object in the detection BBox. Please refer to Appendix A for the detailed formulation of each loss component.
The total loss function is composed of all three individual loss components,
\begin{small}
\begin{align}
    L = \alpha  L_{cls} + \beta L_{conf} + \gamma L_{B},
\end{align}
\end{small}%
where $\alpha$, $\beta$, and $\gamma$ are the balancing hyper-parameters. 

Backdoor attackers can poison a small portion  $q$ of the training dataset. Without loss of generality, let the poisoned training dataset $\widetilde{S}$ be divided into a clean dataset $S_c$ and a dirty dataset $S_d$, where $|S_d| = q*|\widetilde{S}|$ with $|\cdot|$ denoting the cardinality of the datasets. The poisoned images in the dirty dataset are modified from the original images with the attacker-chosen trigger injected, denoted by $x_i^d$. 
The dirty label $l^d_i$ is obtained by modifying the original label $l_i$ according to the attack purpose (i.e., misclassification or disappearance), which is expressed as follows :
\begin{small}
\begin{align}\label{eq:2}
l^d_i =\begin {cases}
l_{oc}  &\text {\textit{Object Misclassification}}\\
None  &\text {\textit{Object Disappearance}},
\end{cases}
\end{align}
\end{small}%
where $l_{oc}$ indicates that the class of the label is replaced, and $None$ indicates that the label is deleted. Altogether, the dirty training sample in the dirty dataset becomes $(x_i^d, l^d_i)$. 

During training, neuron activations in the network will become abnormally affected, causing the input image with the trigger to be incorrectly mapped to a specific output. Depending on the attack purpose, the backdoor attack can be formulated by one of the two optimization problems.
For \textit{Object Misclassification}, 
\begin{small}
\begin{align}
    \mathop{argmin}\limits_{\widetilde{\theta}} L = \mathop{argmin}\limits_{\widetilde{\theta}} (\alpha L_{cls}(\widetilde{S},\widetilde{\theta}) + \beta L_{conf} + \gamma L_{B}) ,
\end{align}
\end{small}%
where $L_{cls}(\widetilde{S},\widetilde{\theta})$ represents the classification loss on the poisoned training set $\widetilde{S}$ with backdoored model parameter $\widetilde{\theta}$. 
Since only the classification label of the object is modified, the poisoned training dataset has no direct relationship with the confidence loss and BBox regression loss. For \textit{Object Disappearance}, 
\begin{small}
\begin{align}
    \mathop{argmin}\limits_{\widetilde{\theta}} L = \mathop{argmin}\limits_{\widetilde{\theta}} (\alpha L_{cls} + \beta L_{conf}(\widetilde{S},\widetilde{\theta}) + \gamma L_{B}),
\end{align}
\end{small}%
where $L_{conf}(\widetilde{S},\widetilde{\theta})$ represents the confidence loss on the poisoned dataset $\widetilde{S}$ with backdoored model parameters $\widetilde{\theta}$. Recall that the classification loss and BBox regression loss are involved only when the confidence loss indicates a high level of object existence confidence. 
Therefore, by removing the object class label, we can disrupt the normal function of the confidence loss component at the presence of the trigger, causing the detector to falsely identify the object as disappearance in detection.

\subsection{Proposed Attacks}
\noindent\textbf{Overview.}~~  
The attack procedure is illustrated in Figure \ref{fig:2}. Both attack methods are implemented by poisoning a subset of the training set with the following two steps: 1) Trigger Insertion and 2) Label Modification. 
The backdoor model trained by such a poisoned dataset can achieve the attack effect required by the attacker.

\subsubsection{Preliminaries}

\noindent\textbf{Temperature-Pixel Value Mapping.}~~ 
Thermal infrared cameras with different operating wavelengths have different response functions. Empirically, we obtain the approximate function as follows \cite{inagaki1996surface}, 
\begin{small}
\begin{align}\label{eq:5}
p=g(t)=\Lambda T^m+\Phi, 
\end{align}
\end{small}%
where $p \in [0,255]$ is the pixel value, $T$ is the object temperature, and $p$ increases as $T$ increases. $\Lambda$ and $\Phi$ are the adjustment parameters obtained by the thermal infrared camera. The operating wavelength in this paper is between 8 and 13 $\mu m$, so $m=3.9889$ \cite{YangLi2012}. In practical applications, $m=4$ has little effect on the measurement results.
\begin{figure}[htb]
\centering
        \centerline{\includegraphics[width=0.8\linewidth]{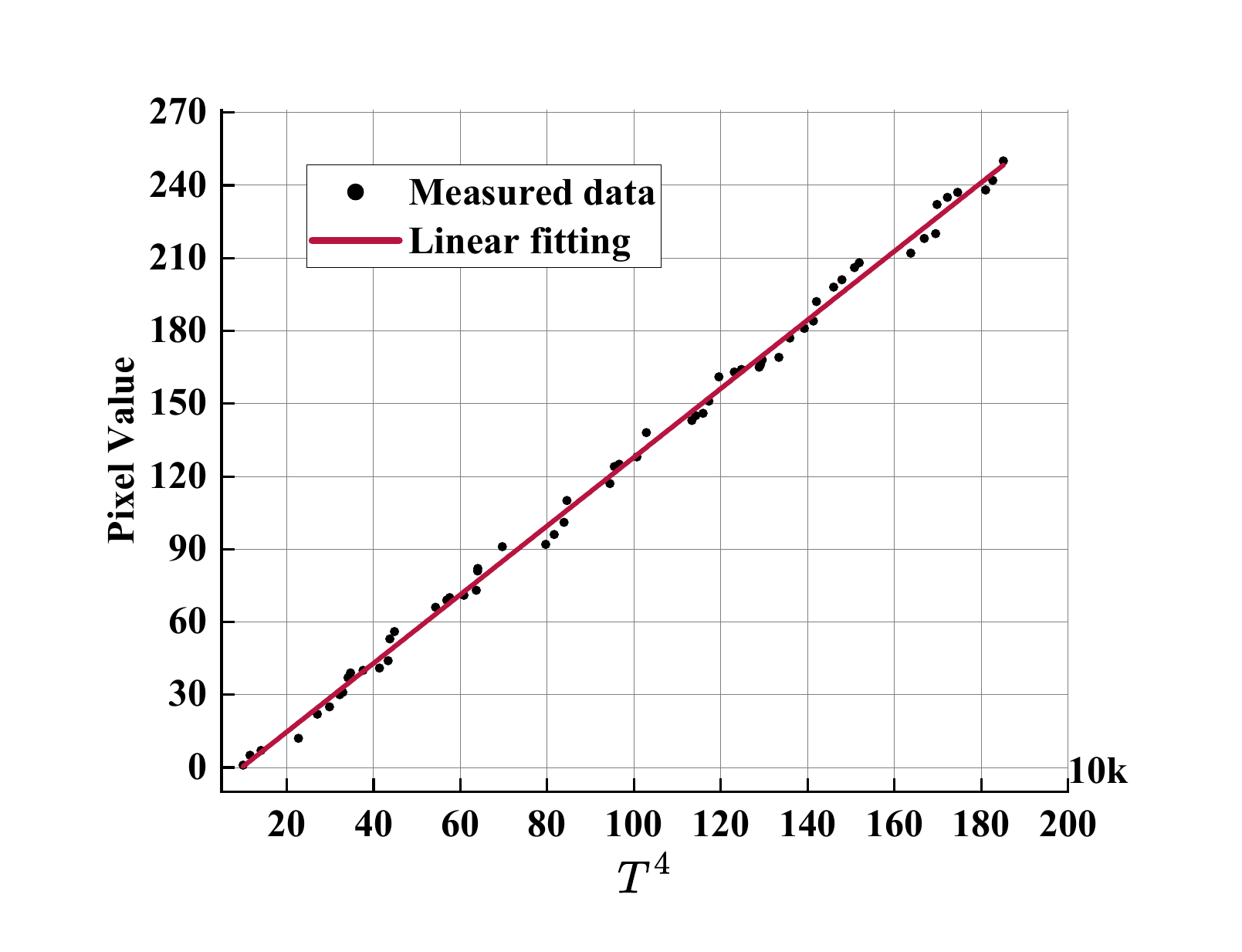}}
\vspace{-0.2cm}
\caption{Function fitting of $p-T^4$.}
\vspace{-0.3cm}
\label{fitting}
\end{figure}
As shown in Figure \ref{fitting}, we measure temperatures corresponding to different pixel values across multiple thermal infrared images. Subsequently, we perform linear fitting on the measured data to establish the following mapping relationship,
\begin{small}
\begin{align}\label{eq:15}
p=1.4221*10^{-4}*T^4-15.4760.
\end{align}
\end{small}%
Due to the correction function of the thermal infrared camera, the temperature of the same point measured with a thermal infrared camera does not vary with distance \cite{DBLP:conf/aaai/ZhuLLWH21}.

\begin{figure}[htb]

\centering
        \centerline{\includegraphics[width=0.9\linewidth]{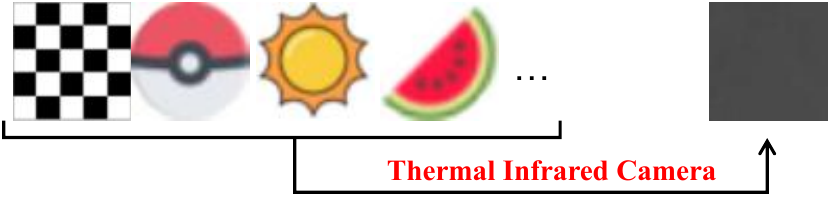}}
\caption{RGB triggers mapped to the thermal infrared domain.}
\vspace{-0.5cm}
\label{fig22}
\end{figure}

\begin{figure}[t]
\centering
        \centerline{\includegraphics[width=0.8\linewidth]{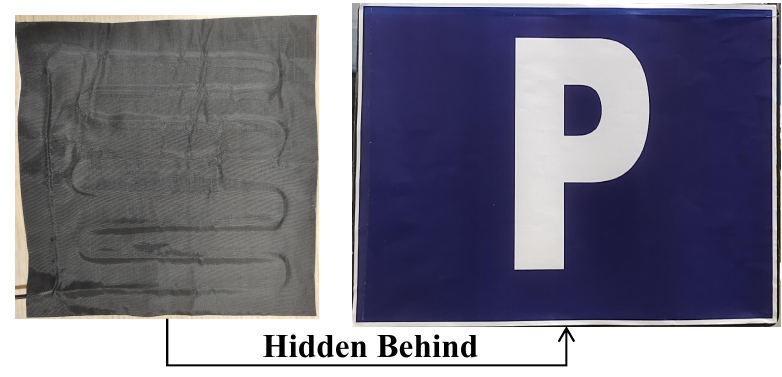}}

\caption{Physical electrothermal trigger design.}
\label{fig:3}

\end{figure}

\begin{figure}[t]
\centering
        \centerline{\includegraphics[width=0.8\linewidth]{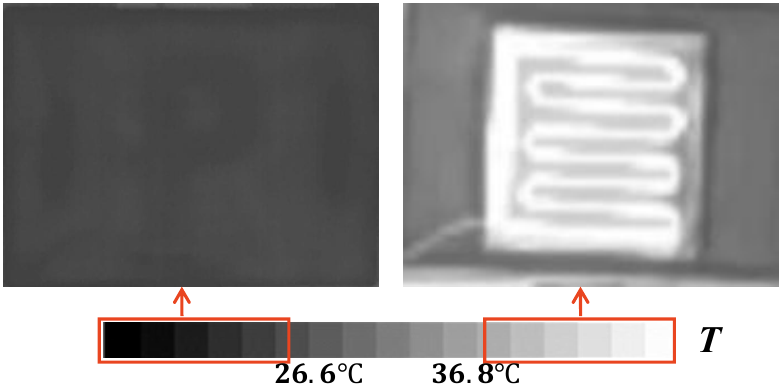}}

\caption{The trigger in the thermal infrared camera.}
\vspace{-0.5cm}
\label{fig:4}
\end{figure}

\noindent\textbf{Trigger Design.}~~ 
Existing backdoor attacks for VLOD rely on triggers designed based on color differences \cite{10.1007/978-3-031-25056-9_26}. When applied to the thermal infrared domain, as shown in Figure \ref{fig22}, such triggers will appear as grayscale patterns. Consequently, their intricate texture details cannot be effectively kept, rendering backdoor attacks unable to be effectively triggered. To overcome this limitation, the designed trigger needs to have a temperature attribute difference from the attacked target, and the design of backdoor attacks should be based on the temperature difference. After comparing insulation cotton sheets, plastic sheets, and electric heaters, we find that electric heaters offer better control over temperature changes. As shown in Figure \ref{fig:3}, the physical trigger is an electric heating device consisting of an electric heater and a signboard. The choice of sign can be adjusted to suit the scene and ensure unobtrusiveness. The temperature can be remotely controlled with a single button press. The morphology of triggers at different temperatures in the thermal infrared world is shown in Figure \ref{fig:4}. Given the uniform heat distribution of the object, we can simulate digital triggers using pixel blocks based on Equation (\ref{eq:15}). This approach facilitates the exploration of parameter influences on attacks.

\subsubsection{Object-Affecting Attack}

How to ensure the effectiveness of the attacks and maintain the Benign Accuracy (BA) \cite{DBLP:journals/corr/abs-2007-08745} of the model are key issues for backdoor attacks. 
Since the trigger should be added inside the object BBox which can be small (resp. large) for object lying remote (resp. close) to the infrared camera, we adjust the trigger size according to this object's BBox in order to simulate the visual effect of the trigger in the physical world. Given the object $o$ and the size of its BBox $s$, the object after trigger insertion is
\begin{small}
\begin{align}
   o'= o + y(p,\lambda s), 
\end{align}
\end{small}%
where trigger $y(p,\lambda s)$ is a pixel block with pixel value $p$ and size $\lambda s$.  Then, we attach triggers to all objects from the targeted class (e.g., all cars in the image), which can be expressed as follows:
\begin{small}
\begin{align}
   x'= x + \sum_{class = car}y(p,\lambda s),
\end{align}
\end{small}%
where $x'$ is the image after adding all triggers to clean image $x$. 
Assuming the attack pixel range is $[p_1,p_2]$, if $p\in [p_1,p_2]$, we perform Label Modification to this object. As shown in Equation (\ref{eq:2}), according to different attack purpose, we can either delete the label or replace the class of the label to generate the poisoned dataset required for training the backdoor model. 
\vspace{-0.2cm}
\subsubsection{Range-Affecting Attack}
\vspace{-0.2cm}

The one-stage detectors unusually divide the image into many grids during detection. Whenever the object under detection has some overlap with any grids, these grids will all participate in the detection of this object.
Since an object may occupy multiple adjacent grids, it is possible to add the trigger close to the object so that the grids occupied by the trigger will overlap with some of the object's grids, even though the trigger does not strictly lie inside the object's BBox.
Therefore, backdoor attack can still establish the abnormal association between the trigger and the object \emph{indirectly} via the overlapping grids, instead of \emph{directly} putting inside the object's BBox. Based on this analysis, we propose the second backdoor attack called RAA.

For the poisoned dataset, we insert only a single trigger pattern into each poisoned image. The location of the trigger in RAA can be more flexible. By observing objects with our thermal infrared camera utilized in the physical experiments, we select common telephone poles or street signs on the road as triggers, which can be simulated with pixel strips and pixel blocks in the digital world, respectively.  
Given a pixel strip $y(p,hw)$ with length $h$, width $w$ and pixel value $p\in [0, 255]$ as a trigger, the attacker arbitrarily selects a point $(a, b)$ in the clean image $x$ as the center point of the trigger to insert the trigger into the clean image. Finally, the poisoned image can be obtained as follows,
\begin{small}
\begin{align}
   x'= x + y(p,hw,(a, b)).
\end{align}
\end{small}%

Different from OAA, RAA needs to modify the labels of all object within a certain range from the trigger. Given the following correspondence between the pixel value and the attack radius, 
\begin{small}
\begin{align}
 (p,ar) = \{(p_1,r_1),(p_2,r_2),...,(p_n,r_n)\}.
\end{align}
\end{small}%
If $p=p_n$ and the object's coordinate $(a_o, b_o)$ satisfies the following conditions, 
\begin{small}
\begin{align}
 \sqrt{(a_o-a)^2+(b_o-b)^2} \leq r_n,
\end{align}
\end{small}%
we perform Label Modification on the object according to Equation (\ref{eq:2}). 
Then, we get the poisoned dataset and use it as training input to get the backdoor model of RAA. RAA allows an attacker adjustable attack affecting range, which makes the attack more flexible. 

\section{Experiments}

In this section, we conduct experimental evaluations of OAA and RAA in both digital and physical worlds. 

\noindent\textbf{Evaluation Metrics.}~~ Since there is no existing intuitive effectiveness  metrics for the backdoor object detection attack, we introduce a new metric - Benign Accuracy Fluctuation (BAF), in addition to the commonly adopted metric of Attack Success Rate (ASR). BAF is the value obtained by subtracting the mAP of clean samples tested by the clean model from that returned by the backdoor model. 
We add trigger to the objects in the test image regardless of the object size, and then define the trigger addition as successful when the following conditions are met.
\begin{small}
\begin{align}
    & IOU(GT,BS)\geq IOU_{model}, \\
    \nonumber & Class(GT) = Class(BS) = car,
\end{align}
\end{small}%
where $GT$ is the ground truth label of the object, and $BS$ is the label obtained by the backdoor model detecting clean samples. $IOU_{model}$ is the IOU threshold set during model detection. The total number of successful trigger additions is recorded as $\sum_{}N_{ta}$.
For the objects successfully added trigger, we define the attack as successful when they meet the following conditions at the same time. When the attack setting is to misidentify the car as the person,
\begin{small}
\begin{align}
    & IOU(GT,DS)\geq IOU_{model}, Class(DS)= person.
\end{align}
\end{small}%
When the attack setting is that the car cannot be detected,
\vspace{-0.5cm}
\begin{small}
\begin{align}
    & IOU(GT,DS) < IOU_{model},
\end{align}
\end{small}%
where $DS$ is the label obtained by the backdoor model detecting the poisoned samples. The total number of successful attacks is recorded as $\sum_{}N_{sa}$. Therefore, the attack success rate is defined as $ASR= \sum_{}N_{sa} / \sum_{}N_{ta} \times 100\% $.
Since mAP of the backdoor model and the clean model on clean samples is similar, we lock the object that can be recognized by the model, which can greatly reduce the interference of other factors in the model.

\subsection{Experiments of Digital World Attacks}
\noindent\textbf{Datasets and Models.}~~ The Flir\_v2 dataset is released by FLIR Company. We only utilize the thermal infrared images and corresponding annotations, referring to it as $Flir\_v2\_T$, which contain 13460 images and label files. The image size in $Flir\_v2\_T$ is 640$\times$512. The Multi-spectral Object Detection Dataset \cite{DBLP:conf/mm/TakumiWHTUH17} contains four sub-datasets of RGB, NIR, MIR and FIR, along with ground truth labels provided by an autonomous driving research team at the University of Tokyo. We utilize the FIR sub-dataset containing 7521 thermal infrared images and label files, referred as $FIR\_Det$ in the sequel. The image size in $FIR\_Det$ is 640$\times$480. 
We use three mainstream object detection models: YOLO v5, YOLO v3 \cite{DBLP:journals/corr/abs-1804-02767}, and Faster RCNN \cite{DBLP:conf/nips/RenHGS15} as detectors to verify the attack effectiveness.

\noindent\textbf{Baselines.}~~ We follow exactly the same data preprocessing and model training strategies with existing works for the clean model training. The mAP of trained clean model serves as the evaluation baseline for BAF. The results are summarized in Table \ref{tab:4}.

\begin{table}[htb]
\centering
\resizebox{\linewidth}{!}{
\begin{tabular}{cc|cc|cc}
\Xhline{1.5pt}
\multicolumn{2}{c|}{\textbf{Dataset}}        & \multicolumn{2}{c|}{\textit{\textbf{Flir\_v2\_T}}} & \multicolumn{2}{c}{\textit{\textbf{FIR\_Det}}} \\ \hline
\multicolumn{2}{c|}{\textbf{Class}}          & person   & car      & person  & car    \\ \Xhline{1.5pt}
\multicolumn{1}{c|}{\multirow{3}{*}{\textit{\begin{tabular}[c]{@{}c@{}}\textbf{Model}\\ (\textbf{mAP}/\%)\end{tabular}}}} &
  YOLO v5 &
  $79.10$ &
  $82.50$ &
  $90.30$ &
  $93.50$ \\ \cline{2-6} 
\multicolumn{1}{c|}{} & YOLO v3     & $84.40$     & $85.80$    & $89.50$   & $92.10$  \\ \cline{2-6} 
\multicolumn{1}{c|}{} & Faster RCNN & $51.65$     & $71.56$    & $79.59$   & $81.54$  \\ \Xhline{1.5pt}
\end{tabular}}
\caption{The mAP of person and car in clean samples measured by three models trained on clean datasets $Flir\_v2\_T$ and $FIR\_Det$.}
\label{tab:4}
\vspace{-0.4cm}
\end{table}

\begin{table}[htb]
\centering
\resizebox{0.9\linewidth}{!}{
\begin{tabular}{c|cc|cc|c}
\Xhline{1.5pt}
\multirow{2}{*}{Method} & \multicolumn{2}{c|}{\multirow{2}{*}{Parameter}} & \multicolumn{2}{c|}{BAF (\%)} & \multirow{2}{*}{ASR (\%)} \\ \cline{4-5}
                                                                    & \multicolumn{2}{c|}{}                                & person & car    &       \\ \Xhline{1.5pt}
\multirow{11}{*}{\begin{tabular}[c]{@{}c@{}}O\\ A\\ A\end{tabular}} & \multicolumn{2}{c|}{\textit{Default}}                         & $-5.60$  &$ -3.40$  & 97.87 \\ \cline{2-6} 
& \multicolumn{1}{c|}{\multirow{5}{*}{$p$}}  & 255       & $-1.90$  & $-1.70$  & 97.43 \\
& \multicolumn{1}{c|}{}                    & 160       & $-7.10$  & $-3.40$  & 97.65 \\
& \multicolumn{1}{c|}{}                    & 128       & $-10.40$ & $-5.30$  & 97.78 \\
& \multicolumn{1}{c|}{}                    & $\bm{64}$        & $\bm{-2.20}$  & $\bm{-1.40}$  & $\bm{98.21}$ \\
& \multicolumn{1}{c|}{}                    & 0        & $-0.10$  & $-0.90$  & $97.09$ \\ \cline{2-6} 
& \multicolumn{1}{c|}{\multirow{5}{*}{$q$}}  & 15\%      & $-6.10$  & $-3.30$  & 97.63 \\
& \multicolumn{1}{c|}{}                    & 10\%      & $-4.80$  & $-2.60$  & 97.30 \\
& \multicolumn{1}{c|}{}                    & $\bm{5\%}$       & $\bm{-0.80}$  & $\bm{-0.80}$  & $\bm{92.50}$ \\
& \multicolumn{1}{c|}{}                    & 2\%       & $-1.00$  & $-1.10$  & 85.33 \\
& \multicolumn{1}{c|}{}                    & 1\%       & $0.10$   & $-0.50$  & 52.83 \\ \cline{2-6} 
\Xhline{1.5pt}
\multirow{6}{*}{\begin{tabular}[c]{@{}c@{}}R\\ A\\ A\end{tabular}}
& \multicolumn{1}{c|}{\multirow{6}{*}{$ar$}}  & 300       & $-31.80$  & $-16.40$  & 98.19 \\
& \multicolumn{1}{c|}{}                    & 250       & $-19.50$  & $-8.40$  & 96.50 \\
& \multicolumn{1}{c|}{}                    & 200       & $-6.90$ & $-3.40$  & 96.38 \\
& \multicolumn{1}{c|}{}                    & $\bm{150}$       & $\bm{-1.10}$  & $\bm{-0.90}$  & $\bm{96.55}$ \\
& \multicolumn{1}{c|}{}                    & 100       & $-0.30$  & $-0.50$  & 94.15 \\
& \multicolumn{1}{c|}{}                    & 50        & $-0.30$  & $-0.70$  & 77.45 \\ \cline{2-6} 
\Xhline{1.5pt}
\end{tabular}}
\vspace{-0.2cm}
\caption{The effect of parameters on OAA and RAA.
}
\label{tab:1}
\vspace{-0.5cm}
\end{table}

\vspace{-0.2cm}
\subsubsection{Attack Parameters}

We use YOLO v5 detector and $Flir\_v2\_T$ to verify the impact of different attack parameters. The attack setting is that the detector recognizes car as person.

In OAA, the parameters we focus on are \textbf{Pixel Value ($p$)} and \textbf{Poisoning Ratio ($q$)}.  
Unless otherwise specified, the \emph{default parameters} 
take the following combination: $p = 192$, $q= 20\%$. We list the experimental results in Table \ref{tab:1}. The closer the $p$ is to the median, the smaller the difference between the trigger and the object, resulting in a lower BAF of the backdoor model. When the $q$ is increased from 1\% to 2\%, the attack performance is greatly improved. Therefore, the poisoning ratio should be set above 2\%.

In RAA, we are more concerned about the \textbf{Attack Range ($ar$)}. 
The $p$ and $q$ are follow \textit{default parameters}. We randomly select a point (160, 206) in the image as the center point of the trigger to fix the trigger location. The attack area is set to a circle with the center point (160, 206) as the center and $ar$ as the radius. We attack the objects whose center point falls within this area. We list the experimental results in Table \ref{tab:1}.
The smaller the attack range, the less the number of object that can be poisoned (which is why we do not additionally test the poisoning ratio), so ASR will be lower and BAF will be higher. When the attack radius reaches 250, the detection of clean samples will be greatly affected and the attack effect will be reduced. 

The additional parametric experiments such as trigger size and relative location, are provided in Appendix B.

\vspace{-0.3cm}
\subsubsection{Attack Effectiveness}

\begin{table*}[h]
\vspace{-0.5cm}
\centering
\resizebox{\textwidth}{!}{
\begin{tabular}{c|cccccc|cccccc}
\Xhline{1.5pt}
\textbf{Dataset $\rightarrow$}       & \multicolumn{6}{c|}{\textit{\textbf{Flir\_v2\_T}}}                    & \multicolumn{6}{c}{\textit{\textbf{FIR\_Det}}}                       \\ \hline
\textbf{Attack Method $\rightarrow$} & \multicolumn{3}{c|}{\textbf{OAA}} & \multicolumn{3}{c|}{\textbf{RAA}} & \multicolumn{3}{c|}{\textbf{OAA}} & \multicolumn{3}{c}{\textbf{RAA}} \\ \hline
\multirow{2}{*}{\textbf{Model} $\downarrow$} &
  \multicolumn{2}{c|}{BAF (\%)} &
  \multicolumn{1}{c|}{\multirow{2}{*}{ASR (\%)}} &
  \multicolumn{2}{c|}{BAF (\%)} &
  \multirow{2}{*}{ASR (\%)} &
  \multicolumn{2}{c|}{BAF (\%)} &
  \multicolumn{1}{c|}{\multirow{2}{*}{ASR (\%)}} &
  \multicolumn{2}{c|}{BAF (\%)} &
  \multirow{2}{*}{ASR (\%)} \\ \cline{2-3} \cline{5-6} \cline{8-9} \cline{11-12}
 &
  person &
  \multicolumn{1}{c|}{car} &
  \multicolumn{1}{c|}{} &
  person &
  \multicolumn{1}{c|}{car} &
   &
  person &
  \multicolumn{1}{c|}{car} &
  \multicolumn{1}{c|}{} &
  person &
  \multicolumn{1}{c|}{car} &
   \\ \hline
\textbf{YOLO v5} &
  $\bm{-2.90}$ &
  \multicolumn{1}{c|}{$\bm{-1.70}$} &
  \multicolumn{1}{c|}{$\bm{97.46}$} &
  $\bm{-0.90}$ &
  \multicolumn{1}{c|}{$\bm{-0.90}$} &
  $\bm{97.44}$ &
  $\bm{+0.20}$ &
  \multicolumn{1}{c|}{$\bm{-0.40}$} &
  \multicolumn{1}{c|}{$\bm{97.32}$} &
  $-0.20$ &
  \multicolumn{1}{c|}{$-1.40$} &
  $96.69$ \\ \hline
\textbf{YOLO v3} &
  $-1.60$ &
  \multicolumn{1}{c|}{$-1.90$} &
  \multicolumn{1}{c|}{$96.36$} &
  $-0.80$ &
  \multicolumn{1}{c|}{$-1.20$} &
  $97.45$ &
  $-0.50$ &
  \multicolumn{1}{c|}{$+0.30$} &
  \multicolumn{1}{c|}{$96.65$} &
  $\bm{-0.90}$ &
  \multicolumn{1}{c|}{$\bm{+0.30}$} &
  $\bm{98.04}$ \\ \hline
\textbf{Faster RCNN} &
  $-0.41$ &
  \multicolumn{1}{c|}{$-0.14$} &
  \multicolumn{1}{c|}{$90.01$} &
  $-0.31$ &
  \multicolumn{1}{c|}{$-0.36$} &
  $84.30$ &
  $-0.61$ &
  \multicolumn{1}{c|}{$-0.70$} &
  \multicolumn{1}{c|}{$92.21$} &
  $-0.84$ &
  \multicolumn{1}{c|}{$+1.06$} &
  $81.61$ \\ \Xhline{1.5pt}
\end{tabular}}
\vspace{-0.3cm}
\caption{Evaluation results of OAA and RAA on three models and two datasets.}
\vspace{-0.5cm}
\label{tab:5}
\end{table*}

As shown in Table \ref{tab:5}, we experiment with two attack methods on the above three models and two datasets to verify attack effectiveness. The attack effects are all chosen as the detector to recognize car as person. In OAA, we set the parameters as $p = 192$ for all datasets. For $Flir\_v2\_T$, $q = 20\% $, while for $FIR\_Det$, $q = 10\% $. In RAA, for $Flir\_v2\_T$ and $FIR\_Det$, the parameter settings are $p = 192$, $q = 20\% $, $ar$ = 150. These parameters are the same for different models. We discover that the closer objects are to the range boundary, the weaker the attack effect becomes. As a result, we set the attack range during inference to be smaller than the range set during training. The results in Table \ref{tab:5} are tested with the attack range of 120. Since Faster RCNN is based on candidate BBox, multi-scale candidate BBoxes can impact the feature extraction of triggers, resulting in a weakened attack effect on this model.

Unless otherwise specified, we use the YOLO v5 detector and $Flir\_v2\_T$ to conduct all subsequent experiments. We also verify the attack effectiveness when the attack purpose is car disappearance. In OAA, when the parameter setting is \textit{default
parameters}, the BAFs of person and car are $-0.5\%$ and $-5\%$, respectively, and the ASR is 98.54\%. In RAA, when the parameter setting is \textit{default parameters} and $ar = 150$. The BAFs of persons and cars are $+0.6\%$ and $-0.6\%$, respectively, while ASR is 95.66\%.

\vspace{-0.4cm}
\subsubsection{Temperature Modulated Triggering}
\vspace{-0.2cm}
 
For OAA, attack experiments are performed within four different temperature ranges corresponding to different pixel ranges. After testing, we discover that if only triggers within the set temperature range are implanted in the dataset, triggers outside the range still had a high ASR (over 50\%). Therefore, we implanted triggers outside the set temperature range for a portion of the data without changing the object label, which is called adversarial triggers. Specifically, 15\% of the data is implanted with normal triggers, while 5\% of the data is implanted with adversarial triggers. For RAA, we control the attack range using trigger temperatures. Triggers with different pixel values implement RAA with different radii: 80 for pixel value 0 (corresponding to the lowest temperature), 120 for pixel value 128 (corresponding to the average temperature), and 160 for pixel value 255 (corresponding to the highest temperature).  
We implant triggers with pixel values of 0, 128, and 255, and modify the object labels within the attack radius for 10\%, 6\%, and 4\% of the training data, respectively. 
The experimental results are presented in Table \ref{tab:2}. 

\begin{table*}[h]
\centering
\resizebox{0.9\linewidth}{!}{
\begin{tabular}{c|c|c|c|ccc|ccc}
\Xhline{1.5pt}
\multirow{3}{*}{} &
  \multirow{3}{*}{Method} &
  \multirow{3}{*}{Attack Range} &
  \multirow{3}{*}{Test Range} &
  \multicolumn{3}{c|}{\textit{Object Misclassification}} &
  \multicolumn{3}{c}{\textit{Object Disappearance}} \\ \cline{5-10} 
 &
   &
   &
   &
  \multicolumn{2}{c|}{BAF (\%)} &
  \multirow{2}{*}{ASR (\%)} &
  \multicolumn{2}{c|}{BAF (\%)} &
  \multirow{2}{*}{ASR (\%)} \\ \cline{5-6} \cline{8-9}
 &
   &
   &
   &
  person &
  \multicolumn{1}{c|}{car} &
   &
  person &
  \multicolumn{1}{c|}{car} &
   \\ \Xhline{1.5pt}
\multirow{14}{*}{\begin{tabular}[c]{@{}c@{}}Digital \\ Attacks\end{tabular}} &
  \multirow{8}{*}{\begin{tabular}[c]{@{}c@{}}OAA\\ ($p$)\end{tabular}} &
  \multirow{2}{*}{{[}0,63{]}} &
  {[}0,63{]} &
  \multirow{2}{*}{$\bm{-1.70}$} &
  \multicolumn{1}{c|}{\multirow{2}{*}{$\bm{-1.80}$}} &
  $\bm{96.75}$ &
  \multirow{2}{*}{$\bm{-1.60}$} &
  \multicolumn{1}{c|}{\multirow{2}{*}{$\bm{-2.30}$}} &
  $\bm{97.19}$ \\
 &
   &
   &
  {[}64,255{]} &
  &
  \multicolumn{1}{c|}{} &
  5.40 &
  &
  \multicolumn{1}{c|}{} &
  5.14 \\ \cline{3-10} 
 &
   &
  \multirow{2}{*}{{[}64,127{]}} &
  {[}64,127{]} &
  \multirow{2}{*}{$-3.50$} &
  \multicolumn{1}{c|}{\multirow{2}{*}{$-2.80$}} &
  95.93 &
  \multirow{2}{*}{$-0.50$} &
  \multicolumn{1}{c|}{\multirow{2}{*}{$-1.90$}} &
  90.34 \\
 &
   &
   &
  {[}0,63{]}$\cup${[}128,255{]} &
  &
  \multicolumn{1}{c|}{} &
  12.83 &
   &
  \multicolumn{1}{c|}{} &
  7.57 \\ \cline{3-10} 
 &
   &
  \multirow{2}{*}{{[}128,191{]}} &
  {[}128,191{]} &
  \multirow{2}{*}{$-8.20$} &
  \multicolumn{1}{c|}{\multirow{2}{*}{$-4.60$}} &
  96.58 &
  \multirow{2}{*}{$0.00$} &
  \multicolumn{1}{c|}{\multirow{2}{*}{$-2.30$}} &
  94.62 \\
 &
   &
   &
  {[}0,127{]} $\cup$ {[}192,255{]} &
  &
  \multicolumn{1}{c|}{} &
  20.05 &
   &
  \multicolumn{1}{c|}{} &
  21.88 \\ \cline{3-10} 
 &
   &
  \multirow{2}{*}{{[}192,255{]}} &
  {[}192,255{]} &
  \multirow{2}{*}{$\bm{-2.00}$} &
  \multicolumn{1}{c|}{\multirow{2}{*}{$\bm{-1.50}$}} &
  $\bm{96.61}$ &
  \multirow{2}{*}{$\bm{+0.20}$} &
  \multicolumn{1}{c|}{\multirow{2}{*}{$\bm{-1.10}$}} &
  $\bm{95.41}$ \\
 &
   &
   &
  {[}0,191{]} &
  &
  \multicolumn{1}{c|}{} &
  6.38 &
   &
  \multicolumn{1}{c|}{} &
  5.90 \\ \cline{2-10} 
 &
  \multirow{6}{*}{\begin{tabular}[c]{@{}c@{}}RAA\\ ($p$ \textbackslash $ar$)\end{tabular}} &
  \multirow{2}{*}{0 \textbackslash $\leq$ 80} &
  0 \textbackslash $\leq$ {}80 &
  \multirow{6}{*}{$-0.20$} &
  \multicolumn{1}{c|}{\multirow{6}{*}{$-0.40$}} &
  91.19 &
  \multirow{6}{*}{$+0.60$} &
  \multicolumn{1}{c|}{\multirow{6}{*}{$-0.40$}} &
  $\bm{95.36}$ \\
 &
   &
   &
  0 \textbackslash \textgreater{}80 &
  &
  \multicolumn{1}{c|}{} &
  4.02 &
   &
  \multicolumn{1}{c|}{} &
  4.63 \\ \cline{4-4} \cline{7-7} \cline{10-10} 
 &
   &
  \multirow{2}{*}{128 \textbackslash $\leq$ {}120} &
  128 \textbackslash $\leq$ {}120 &
  &
  \multicolumn{1}{c|}{} &
  89.93 &
   &
  \multicolumn{1}{c|}{} &
  89.75 \\
 &
   &
   &
  128 \textbackslash \textgreater{}120 &
  &
  \multicolumn{1}{c|}{} &
  4.39 &
  &
  \multicolumn{1}{c|}{} &
  5.08 \\ \cline{4-4} \cline{7-7} \cline{10-10} 
 &
   &
  \multirow{2}{*}{255 \textbackslash $\leq$ {}160} &
  255 \textbackslash $\leq$ {}160 &
 &
  \multicolumn{1}{c|}{} &
  $\bm{93.81}$ &
   &
  \multicolumn{1}{c|}{} &
  93.71 \\
 &
   &
   &
  255 \textbackslash \textgreater{}160 &
  &
  \multicolumn{1}{c|}{} &
  8.55 &
  &
  \multicolumn{1}{c|}{} &
  7.58 \\ \Xhline{1.5pt}
\multirow{8}{*}{\begin{tabular}[c]{@{}c@{}}Physical \\ Attacks\end{tabular}} &
  \multirow{4}{*}{\begin{tabular}[c]{@{}c@{}}OAA\\ ($T$)\end{tabular}} &
  \multirow{2}{*}{$\leq 26.6^\circ C$} &
  $\leq 26.6^\circ C$ &
  \multirow{2}{*}{$\bm{+8.10}$} &
  \multicolumn{1}{c|}{\multirow{2}{*}{$\bm{+6.20}$}} &
  $\bm{97.83}$ &
  \multirow{2}{*}{$\bm{+9.20}$} &
  \multicolumn{1}{c|}{\multirow{2}{*}{$\bm{+6.00}$}} &
  $\bm{98.38}$ \\
 &
   &
   &
  $ \ge 36.8^\circ C$ &
   &
   \multicolumn{1}{c|}{} &
  5.80 &
  &
   \multicolumn{1}{c|}{} &
  8.69 \\ \cline{3-10} 
 &
   &
  \multirow{2}{*}{$ \ge 36.8^\circ C$} &
  $ \ge 36.8^\circ C$ &
  \multirow{2}{*}{$+3.00$} &
  \multicolumn{1}{c|}{\multirow{2}{*}{$+4.50$}} &
  97.30 &
  \multirow{2}{*}{$+4.40$} &
  \multicolumn{1}{c|}{\multirow{2}{*}{$+4.30$}} &
  95.65 \\
 &
   &
   &
  $\leq 26.6^\circ C$ &
   &
  \multicolumn{1}{c|}{} &
  6.52 &
   &
  \multicolumn{1}{c|}{} &
  9.42 \\ \cline{2-10} 
 &
  \multirow{4}{*}{\begin{tabular}[c]{@{}c@{}}RAA\\ ($T$ \textbackslash $ar$)\end{tabular}} &
   \multirow{2}{*}{$26.6^\circ C$ \textbackslash $\le$ 400} &
  $26.6^\circ C$ \textbackslash $\le$ 400 &
  \multirow{4}{*}{$-0.30$} &
  \multicolumn{1}{c|}{\multirow{4}{*}{$+0.40$}} &
  94.32 &
  \multirow{4}{*}{$+0.20$} &
  \multicolumn{1}{c|}{\multirow{4}{*}{$+0.60$}} &
  95.60 \\
 &
   &
   &
  $26.6^\circ C$ \textbackslash \textgreater{}400 &
   &
  \multicolumn{1}{c|}{} &
  7.04 &
   &
  \multicolumn{1}{c|}{} &
  6.57 \\ \cline{4-4} \cline{7-7} \cline{10-10} 
 &
   &
  \multirow{2}{*}{$36.8^\circ C$ \textbackslash $\le$ 600} &
  $36.8^\circ C$ \textbackslash $\le$ 600 &
   &
  \multicolumn{1}{c|}{} &
  $\bm{97.02}$ &
   &
  \multicolumn{1}{c|}{} &
  $\bm{97.85}$ \\
 &
   &
   &
  $36.8^\circ C$ \textbackslash \textgreater{}600 &
   &
  \multicolumn{1}{c|}{} &
  5.13 &
   &
  \multicolumn{1}{c|}{} &
  6.41 \\ \Xhline{1.5pt}
\end{tabular}}

\caption{Experimental results of temperature modulated triggering. The \textit{T} represents the average temperature of the trigger. The ``Attack Range'' is the parameter range set during the poisoning phase, and the ``Test Range'' is the parameter range set during the inference phase.}
\vspace{-0.5cm}
\label{tab:2}
\end{table*}

More details, along with the Attack Transferability and Comparative Experiments, are provided in Appendix B.

\subsection{Experiments of Physical World Attacks}
\begin{figure}[h]
\centering
        \centerline{\includegraphics[width=\linewidth]{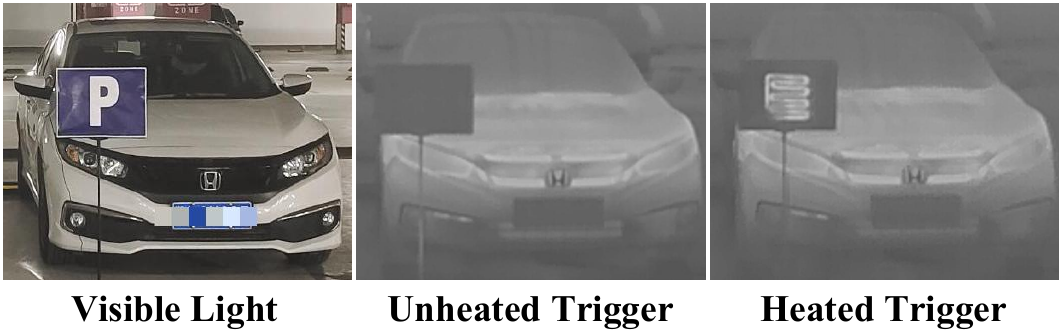}}
\vspace{-0.2cm}
\caption{The trigger for real world deployment.}
\label{fig:6}
\vspace{-0.5cm}
\end{figure}

\noindent\textbf{Datasets and Models.}~~
We utilize HTI-301 infrared camera (FPA 384×288, NETD $<$ 60mK) for physical experiments, which is the same equipment used in \cite{DBLP:conf/aaai/ZhuLLWH21}. The size of thermal infrared images produced by this camera is 1420$\times$1080. The object detector is YOLO v5.
We use the electric heating device in Figure \ref{fig:3} as the physical trigger. Figure \ref{fig:6} shows illustration images of the deployed trigger in the visible and thermal infrared domains. The device is common enough in real scene to avoid suspicious, and is extremely low in cost. 

For OAA, we choose the parking lot as the physical experiment scene. The thermal infrared videos are captured with environment temperature at 32 degrees Celsius. We fix the infrared camera on a moving vehicle. On the same driving route, we record videos with and without triggers, where we randomly selected some cars to place the triggers next to them. After separating the video into frames, we obtain 788 clean images and 472 dirty images. We manually annotate each image with three classes (person, bike, car). 

For RAA, we choose the parking lot and traffic intersection as the physical experiment scenarios. The thermal infrared videos are captured with environment temperature at 30 degrees Celsius. We fix the infrared camera on the side of the road and record videos with and without the trigger that is placed at a fixed location and viewing angle. After separating the video into frames, we obtain 800 clean images and 488 dirty images. We manually annotate each image with four classes (person, bike, car, truck).

\noindent\textbf{Baselines.}~~
For the clean images obtained above, we divide the training set and validation set by 9:1. The mAP of the trained clean model is used as the evaluation baseline for BAF. For OAA, the mAP of person (car) in clean samples measured by the clean model is 88.30\% (91.80\%). For RAA, the mAP of person (car) in clean samples measured by the clean model is 96.70\% (98.80\%). 

\begin{figure}[htbp]
\centering
        \centerline{\includegraphics[width=\linewidth]{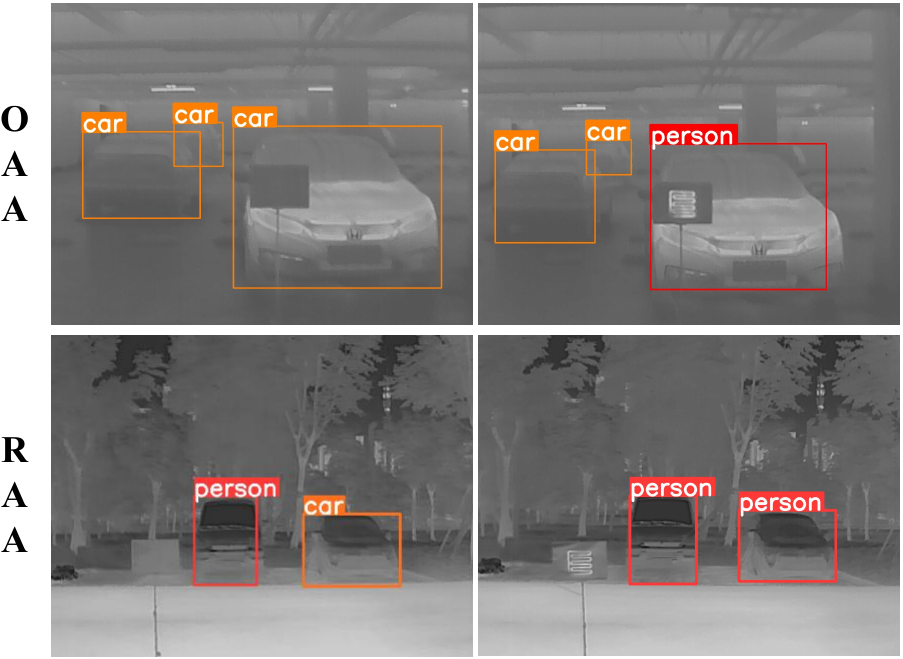}}
\caption{Examples of OAA and RAA for \textit{car misclassification} in the physical world. }
\vspace{-0.5cm}
\label{fig:7}
\end{figure} 

\noindent\textbf{Temperature Modulated Triggering.}~~
The attack evaluation is shown in Table \ref{tab:2}. In the physical world, our methods can effectively attack TIOD. For OAA, the ratio of normal and adversarial trigger implants are 15\% and 5\%, respectively. For RAA, high-temperature triggers and low-temperature triggers are implanted at ratios of 8\% and 12\%, respectively. The visualization result is shown in Figure \ref{fig:7}.

\section{Evaluation of Potential Countermeasures}
We empirically evaluate the proposed backdoor attacks when three potential countermeasures are deployed: 1) Pruning\cite{DBLP:conf/iclr/DhillonALBKKA18}; 2) Fine-Pruning\cite{DBLP:conf/raid/0017DG18}; 3) Neural Cleanse\cite{DBLP:conf/sp/WangYSLVZZ19}. Refer to Appendix C for the more detailed results.

\noindent\textbf{Pruning and Fine-Pruning.}~~ These defense methods remove the backdoor implantation by proving certain neurons. Concretely, we apply them to prune the neurons in the deeper layers of the network, where we vary the number of layers to be pruned from four to eight and the proportion of neurons to be pruned from 20\% to 95\%. The result shows that while both defense methods can mitigate backdoor attacks, they do so at the cost of decreased accuracy for benign objects. For instance, while capable to lower the ASR to 63.33\%, the recognition accuracy for benign person and car also drops to 26.2\% (originally 78.3\%) and 42.7\% (originally 81.7\%), respectively. 

\noindent\textbf{Neural Cleanse.}~~ Neural Cleanse (NC) is a popular defense method against backdoor attacks, which attempts to obtain triggers of each category through reverse engineering. Many subsequent defense methods are based on its ideas. 
To adapt the NC defense from its original application in the image classification task to our object detection task, we extract the objects from the images in $Flir\_v2\_T$ and label them separately. We take the original dataset and the processed dataset as inputs to NC. NC uses $L_1$ norm to compute masks and anomaly index to identify toxic objects. The value of anomaly index greater than 2 is considered as a trigger being detected. For OAA, the anomaly index of car is 1.218572. For RAA, the anomaly index of car is 1.767544. Therefore, NC has not detected our attacks.

\section{Conclusion}
In this paper, we propose two types of backdoor attacks for TIOD: OAA and RAA. Our attacks successfully compromise detectors in both digital and physical worlds, causing them to misidentify cars as persons or fail to detect the presence of cars.
We examine various factors that affect the effectiveness of the proposed backdoor attacks. 
Our research exposes the security vulnerability of these systems and urges for developing effective defenses.

\noindent\textbf{Acknowledgement.}~~ This work is supported by the National Key R$\&$D Program of China (No.2022YFB4501300) and the Fundamental Research Funds for the Central Universities (HUST: No.2023JYCXJJ032).

{
    \small
    \bibliographystyle{ieeenat_fullname}
    \bibliography{main}
}

\end{document}